\pdfoutput=1

\RequirePackage{amsmath}

\documentclass[letterpaper, 10 pt, conference]{ieeeconf}
\IEEEoverridecommandlockouts
\let\labelindent\relax

\makeatletter
\let\NAT@parse\undefined
\makeatother                               

\usepackage[per=frac,binary-units=true]{siunitx}
\usepackage{graphicx}
\usepackage[caption=false, font=footnotesize]{subfig}
\usepackage{bm}
\usepackage{amssymb}
\usepackage{amsfonts}
\usepackage{enumitem} 
\usepackage{multirow}
\usepackage{booktabs}
\usepackage{algorithm,algpseudocode}
\usepackage{url} 
\usepackage{xspace}
\usepackage[T1]{fontenc}
\usepackage{units}
\usepackage[utf8]{inputenc}
\usepackage[hidelinks]{hyperref}
\usepackage{cleveref}
\usepackage{balance}

\usepackage{xcolor}

\usepackage{tikz}
\usetikzlibrary{arrows}
\usetikzlibrary{positioning,calc}
\usetikzlibrary{decorations.pathreplacing}
\usetikzlibrary{decorations.markings}
\usetikzlibrary{fit}
\usetikzlibrary{shapes.callouts}
\usetikzlibrary{shapes.geometric}
\usetikzlibrary{matrix}

\usepackage[comma,numbers,compress]{natbib}

\graphicspath{{images/}}
\DeclareGraphicsExtensions{.pdf,.png,.jpg,.jpeg,.JPG}

\title{\LARGE \bf
Autonomous Bimanual Functional Regrasping\\of Novel Object Class Instances}

\author{Dmytro Pavlichenko$^{*}$, Diego Rodriguez$^{*}$, Christian Lenz$^{*}$, Max Schwarz and Sven Behnke
\thanks{\hspace{-2.2ex}$^{*}$: Authors contributed equally. All authors are with the Autonomous Intelligent Systems (AIS) Group, Computer Science Institute VI, University of Bonn, Germany,
        {\tt\small <lastname>@ais.uni-bonn.de}.}
}%
\usepackage{eso-pic}

\AtBeginDocument{\AddToShipoutPictureFG*{\AtTextUpperLeft{\put(0,\LenToUnit{9pt}){\parbox{\textwidth}{\centering\bfseries International Conference on Humanoid Robots (Humanoids), Toronto, Canada, 2019
}}}}}

\begin{document}
\maketitle
\thispagestyle{empty}
\pagestyle{empty}

\begin{abstract}
In human-made scenarios, robots need to be able to fully operate objects in their surroundings,
i.e., objects are required to be functionally grasped rather than only picked.
This imposes very strict constraints on the object pose such that a direct grasp can be performed. 
Inspired by the anthropomorphic nature of humanoid robots, 
we propose an approach that first grasps an object with one hand,
obtaining full control over its pose, 
and performs the functional grasp with the second hand subsequently.
Thus, we develop a fully autonomous pipeline for dual-arm functional regrasping of novel familiar objects, 
i.e., objects never seen before that belong to a known object category, e.g., spray bottles.
This process involves semantic segmentation, object pose estimation, non-rigid mesh registration, grasp sampling, handover pose generation and in-hand pose refinement.
The latter is used to compensate for the unpredictable object movement during the first grasp.
The approach is applied to a human-like upper body.
To the best knowledge of the authors, this is the first system that exhibits autonomous bimanual functional regrasping capabilities. 
We demonstrate that our system yields reliable success rates and can be applied on-line to real-world tasks
using only one off-the-shelf RGB-D sensor.
\end{abstract}

\section{Introduction}
\label{sec:introduction}
Grasping and manipulating \textit{novel} objects is a fundamental skill that robots need to possess in order to work in daily-life scenarios.
In contrast to object picking, where the object usage is usually completely ignored,
objects often need to be grasped in a very specific manner to ensure their usage,
e.g. when grasping a cordless driller for drilling a hole.
This specific way of grasping is characterized as \textit{functional}.
In many situations, however, desired functional grasps cannot be achieved
in a straightforward fashion.
Collisions with the environment, self-collisions and kinematic limitations may impose constraints on reaching the functional grasps.
A natural solution is to use one (supportive) hand to grasp the object, and hand it over to the second hand so that the desired grasp is achieved. 
We refer to this operation as functional regrasping. 
The ability to perform functional regrasping significantly extends the capabilities of any dual-arm humanoid robot.

Inspired by our previous work on bimanual manipulation~\cite{Pavlichenko2018}, 
in this paper we propose a complete autonomous pipeline that tackles the problem of dual-arm \textit{functional regrasping} of previously unseen objects of a known category. 
The system consists of the following components: object recognition and segmentation, pose estimation, non-rigid object registration, functional grasp inference, grasp sampling for the supportive hand, handover pose computation and in-hand pose refinement. 
Functional regrasping is a non-trivial problem, it imposes several challenges in terms of perception and grasp planning.
Small errors of the object or grasp pose may propagate through all components,
that will finally result in an unsuccessful or non-functional grasp.
Moreover, the system needs to be robust against noisy sensory measurements and modeling imperfections that are typically present in unstructured environments.

\begin{figure}
	\centering
	\includegraphics[height=3.7cm]{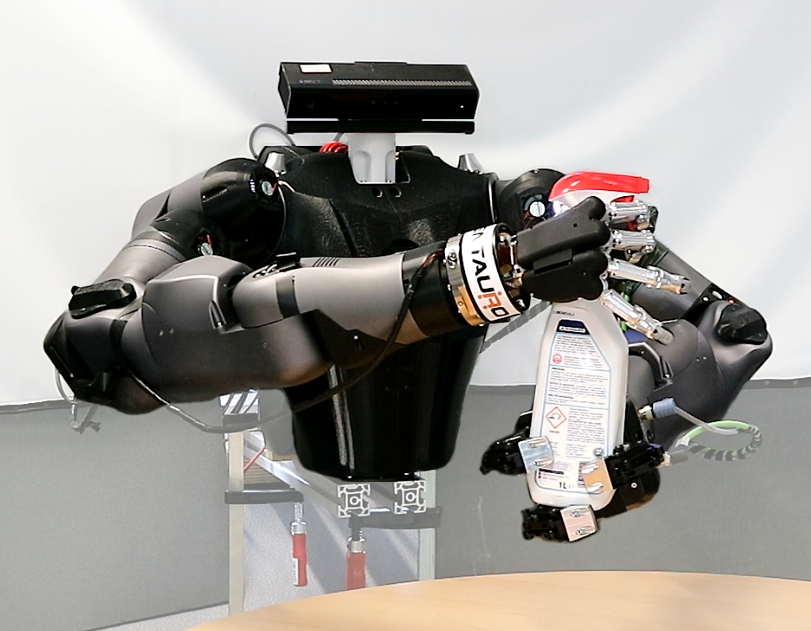}
	\includegraphics[height=3.7cm]{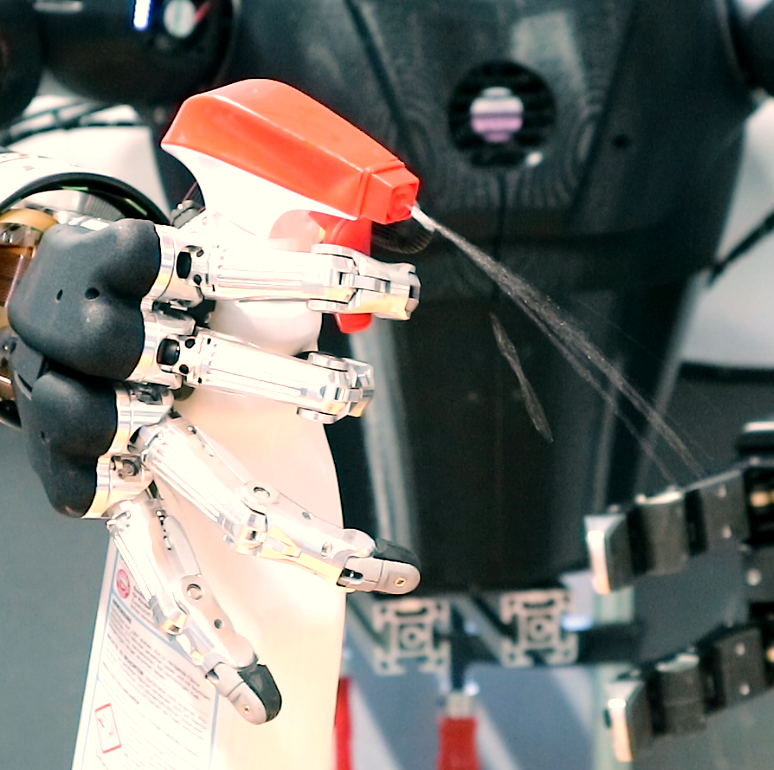}
	\caption{Left: functional regrasp of a spray bottle. Right: functional use of the spray bottle after a successful regrasp.}
	\label{fig:teaser}
	\vspace*{-2ex}
\end{figure}

The main contribution of this paper is the development of a fully autonomous pipeline for dual-arm functional regrasping of previously unseen objects of a known category.
Every component is designed to be maximally independent, which increases the flexibility and upgradeability of the entire pipeline.
Our system is agnostic to the robot kinematics and utilizes only one commercial RGB-D sensor.
These two factors indicate the applicability of our approach to other platforms.
The pipeline is tested on a humanoid torso with two arms equipped with a Robotiq three-finger gripper and a SVH Schunk hand (\cref{fig:teaser}).
During tests six objects from two categories: spray bottles and watering cans were successfully regrasped in a functional way.
The runtime of the entire pipeline takes less than half a minute which makes it suitable for online applications.

\section{Related Work}
\label{sec:related_work}
In contrast to the extensively researched single-arm regrasping strategies~\cite{Terasaki1998,paolini2016,sundaralingam2018,chavan2018}, 
the task of dual-arm regrasping has drawn much less attention in the grasping community,
mainly due to the difficulties caused by the vast number of combinations between two possible grasps~\cite{Koga1992,Koga1994},
and the uncertainty of the object pose after an initial grasp.

\citet{Hou-2018-105030} propose an algorithm for reorienting objects on a table with a two-finger pinch gripper using pivoting and compliant rolling. The exact object meshes are known in advance which allows for an offline precomputation of possible grasps for each object. The method can tolerate slight errors due to the use of simplified meshes. The authors evaluate the approach in simulation and on the real robot. 
Our approach, on the other hand, focuses on achieving a functional grasp and does not require a prior knowledge of the exact shape of the object. 
Instead of relying on gripper precision and correctness of certain physical parameters such as friction of the table surface, we perform a visual pose correction step after the initial grasp is executed.

Dual-arm regrasping approaches have been proposed to extend the workspace of single arm manipulators. 
\citet{Saut2010PlanningPT} present a method for pick-and-place tasks when the initial position of the object lies outside the workspaces of a manipulator.
Offline grasp lists and robot roadmaps are used to reduce the online planning time.
Experiments are performed in simulation with a humanoid robot using ground truth object models.
In contrast, our approach is tested on a real robot and performs functional grasps.

\citet{Weiwei2015} compare single-arm regrasps with dual-arm regrasps.  
For single-arm regrasps, the authors assume a flat surface such as a table in front of the robot which is used to orient the object freely before the target grasp is possible, similar as in~\cite{paolini2016} and~\cite{chavan2018}. 
Grasp graph search is used to find an optimal solution for both regrasp strategies. 
In case of a single regrasp, the stability of the object placement is optimized, while for the dual-arm regrasp manipulability and approachability are maximized.
Although the size of the object is a limitation for dual-arm regrasping, 
single-arm regrasping imposes constraints on the geometry of the objects by demanding a stable placement on a flat surface.

\citet{Vezzani2017} propose a pipeline for bi-manual handover of known objects. 
Tactile and visual feedback are used for object pose estimation and grasp stabilization. 
In-hand localization of the object is performed to increase the precision of the second grasp. 
In order to obtain the handover pose, the known objects are annotated with a set of grasps, reachable by the second hand. 
This pipeline is similar to the one proposed in this paper.
However, our method goes beyond in several aspects:
the objects presented to the system are novel; object-specific annotations are not required but instead grasping poses are transferred to the observed instance; and lastly the objects are functionally grasped.

Recently, \citet{Cruciani2019} have combined in-hand manipulation and dual-arm regrasping.
The authors introduce a Dexterous Manipulation Graph (DMG) to plan the actions for achieving the desired pose of the object in the hand. 
Thus, the problem is formulated as a graph search from the initial to the goal configuration. 
The method is tested on the ABB Yumi robot using parallel grippers on known objects equipped with Apriltags for pose estimation.
On the other hand, our approach handles novel objects and does not require any object modification for estimating the pose, 
due to the use of machine learning approaches for segmentation and pose estimation.

In contrast to previous approaches, \citet{Balaguer2012} address bimanual regrasping of known objects using a model which was previously trained by supervised learning for unimanual grasps. 
Later, an optimizer searches for an actual handover configuration which minimizes the execution time. 
The authors argue that the role of vision in bimanual tasks for humans is low and run a whole manipulation system in an open-loop fashion.
Although their approach may be applicable where the exact grasping pose on the object does not matter, e.g., when using parallel underactuated grippers,
this is not suitable for functional grasps. 
In contrast, our approach incorporates a visual in-hand object pose refinement after the object was grasped for the first time.

To the best of our knowledge there are no recent works that address the problem of dual-arm \textit{functional} regrasping.

\section{System Overview}
\label{sec:system_overview}
\begin{figure}[]
	\centering
	\scalebox{1.0}{\input{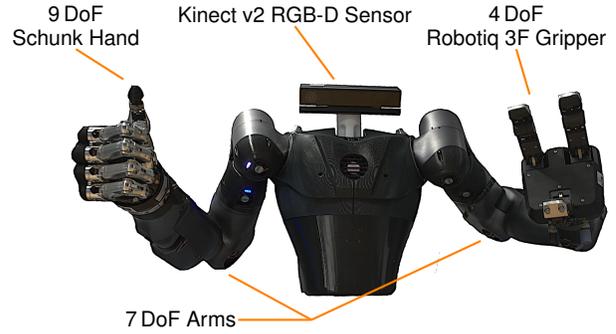}}
	\caption
	{Robot platform used for testing. Due to the anthropomorphic design of the right hand, it is used for the final functional grasp.}
	\label{fig:upperbody}
	\vspace*{-2ex}
\end{figure}

\begin{figure*}[ht!]
	\centering
	\includegraphics[width=\linewidth]{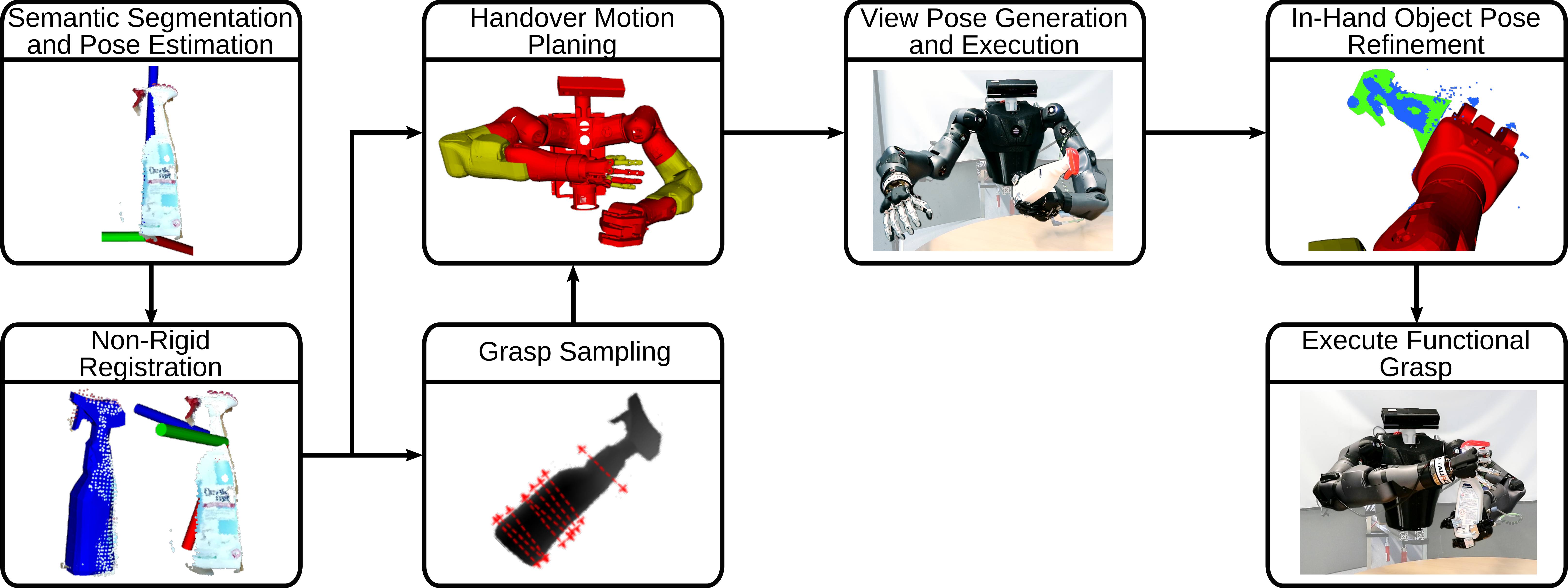}
	\caption{Autonomous regrasping pipeline. First, from the RGB-D data, the object point cloud and its 6D pose are obtained by means of semantic segmentation and class-level pose estimation. 
		Then, a mesh that fits the segmented point cloud is found by using a shape space non-rigid registration.
		The resulting deformation field, that deforms the canonical mesh into the observed instance, allows to warp the functional grasp (predefined for the canonical mesh). 
		The deformed mesh is then used to sample grasp candidates for the supportive arm with Dexterity Network.
		Given the functional grasp and the set of grasp candidates, an optimal handover configuration is found.
		Later, a view pose is generated agnostic to the object. 
		Next, the object is grasped using the supportive hand and the in-hand pose estimation is performed.
		Finally, by means of the object pose correction we refine the initial handover motion and execute it.}
	\label{fig:system_diagram}
	\vspace*{-2ex}
\end{figure*}

We test our software pipeline with a human-like upper body, which is very similar to the upper body of the Centauro robot~\cite{Klamt2018}.
The robot has two anthropomorphic manipulators with 7 Degrees of Freedom (DoF) each. 
The right manipulator is equipped with a Schunk SVH hand with 20 joints and 9~DoF, resulting in 11 mimic joints.
The left manipulator is equipped with a 4~DoF Robotiq three-finger gripper. 
The head of the robot is equipped with a Kinect~v2~\cite{fankhauser2015}.
This is the only sensor used to perform the autonomous functional regrasping tasks. 
The robot platform is showed in \cref{fig:upperbody}.

Our software pipeline~(\cref{fig:system_diagram}) incorporates:
\begin{itemize}
	\item \textit{Semantic Segmentation and 6D Pose Estimation} which outputs a point cloud and a 6D estimate of the object pose using input RGB-D data from the Kinect v2;
	\item \textit{Non-rigid (Shape) Registration} that deforms a canonical model to match the segmented oriented object point cloud and warp the associated functional grasp pose (for the right arm);
	\item \textit{Grasp Sampling} to generate grasp candidates for the supportive (left) arm using as input the deformed mesh;
	\item \textit{Handover Pose Generation} to find the optimal handover configuration considering the functional pose and grasp candidates of the supportive hand;
	\item \textit{Observation Pose Generation} to find a suitable pose for the in-hand object pose refinement; and finally
	\item \textit{In-hand Pose Refinement} to correct the functional grasp pose due to unpredictable displacements during the first grasp.
\end{itemize}

\section{Perception}
\label{sec:perception}
In order to segment the object and separate it from the background,
we train a semantic segmentation network. Additional outputs are added for
estimating the object pose.

\subsection{Training Data}

\begin{figure}[b]
	\centering
	\newlength{\imgwidth}
	\setlength{\imgwidth}{1.8cm}
	\begin{tabular}{ccccc}
		\includegraphics[height=\imgwidth,clip,trim=100 0 30 0]{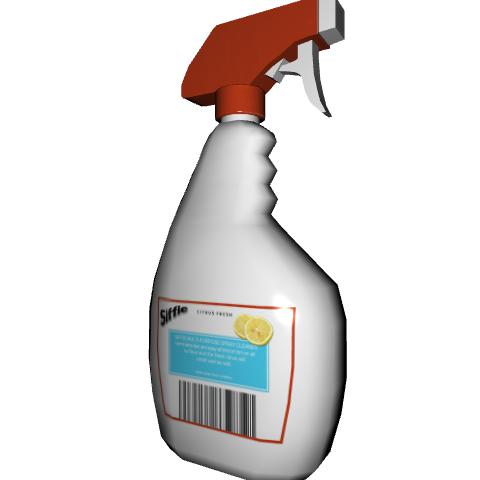} &
		\includegraphics[height=\imgwidth,clip,trim=100 0 30 0]{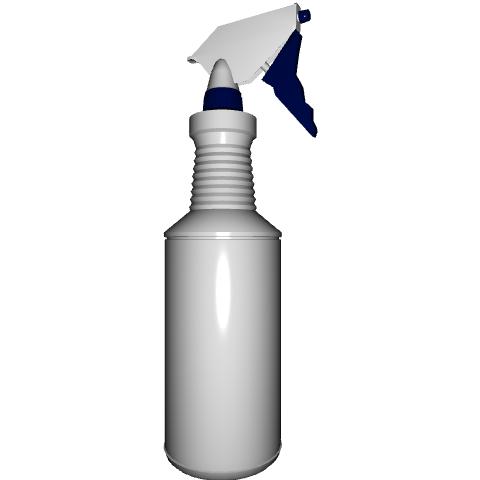} &
		\includegraphics[height=\imgwidth,clip,trim=100 0 30 0]{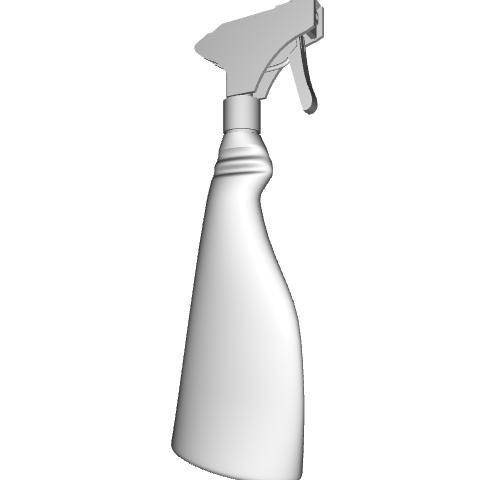} &
		\includegraphics[height=\imgwidth,clip,trim=100 0 30 0]{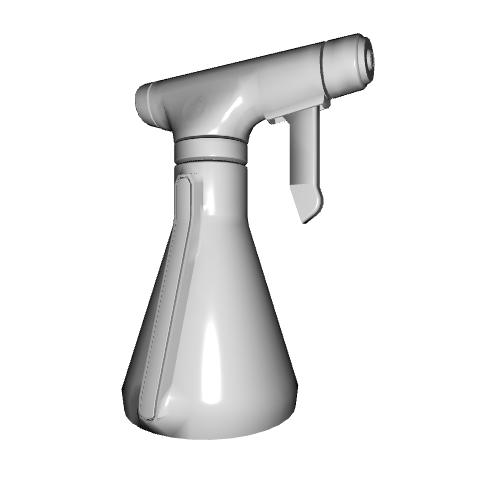} &
		\includegraphics[height=\imgwidth,clip,trim=100 0 30 0]{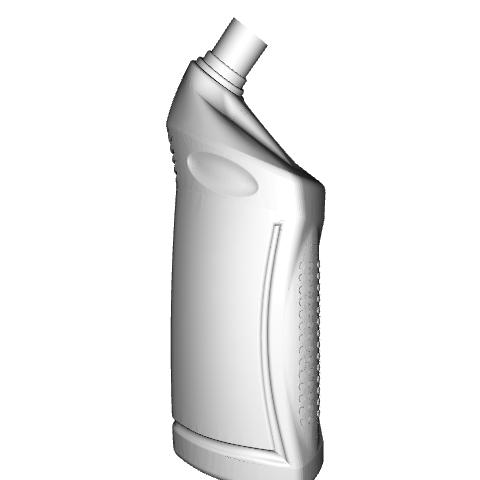} \\
		\includegraphics[height=\imgwidth,clip,trim=100 0 30 0]{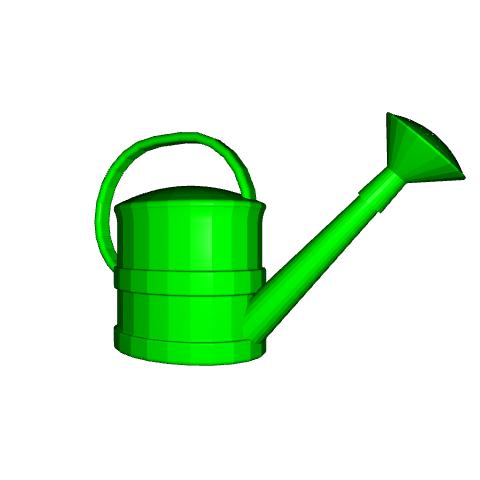} &
		\includegraphics[height=\imgwidth,clip,trim=100 0 30 0]{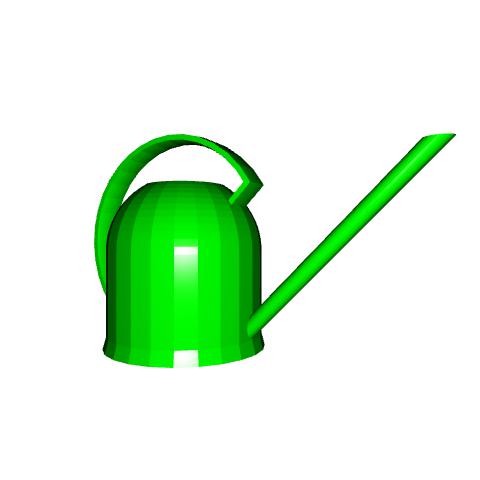} &
		\includegraphics[height=\imgwidth,clip,trim=100 0 30 0]{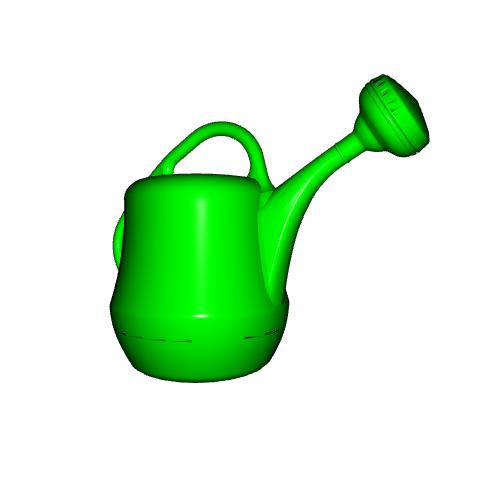} &
		\includegraphics[height=\imgwidth,clip,trim=100 0 30 0]{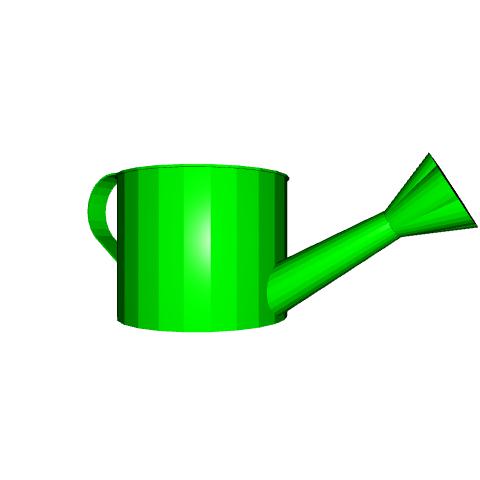} &
		\includegraphics[height=\imgwidth,clip,trim=100 0 30 0]{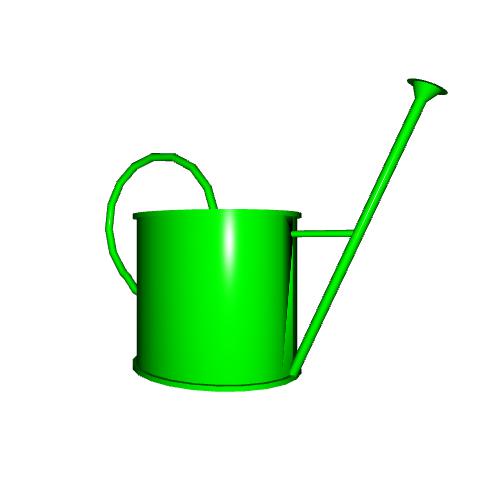} \\
	\end{tabular}
	\caption{Examples of the meshes used for training the segmentation and pose
		estimation model.}
	\label{fig:perception:meshes}
\end{figure}

\begin{figure*}
	\centering
	\includegraphics[height=2.4cm]{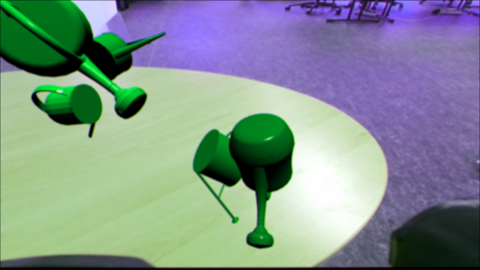}
	\includegraphics[height=2.4cm]{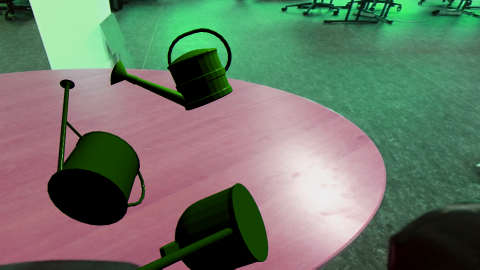}
	\includegraphics[height=2.4cm]{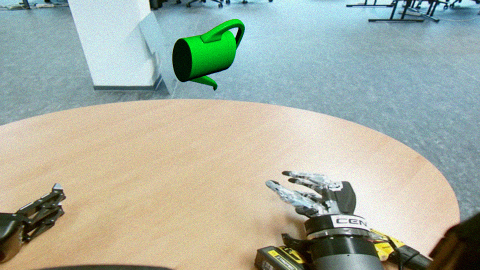}
	\includegraphics[height=2.4cm]{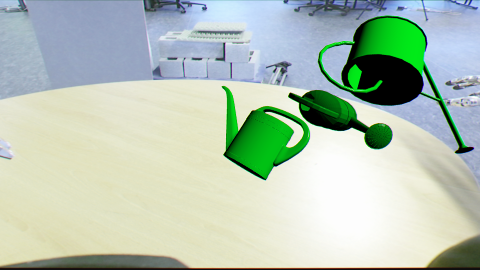}
	\caption{Synthetic training scenes for semantic segmentation and 6D pose estimation.}
	\label{fig:synthetic_scenes}
	\vspace*{-2ex}
\end{figure*}

Deep learning methods need large amounts of training data. Instead of manually
collecting and annotating object images, we follow a hybrid synthetic approach
and render object meshes collected from online databases on top of real
background scenes captured by the robot (\cref{fig:synthetic_scenes}).
\Cref{fig:perception:meshes} shows some of the meshes that make up the training
set---we used seven different spray bottle meshes and nine watering can meshes.
The meshes are manually aligned into a common coordinate system.

\subsection{Semantic Segmentation and Object Pose Estimation}

For semantic segmentation, we implemented the \textit{Lightweight RefineNet}
architecture~\citep{nekrasov2019light} with a ResNet-50 backbone.
In order to estimate a rough 6D pose, we also densely predict a direction to
the projected object center, object depth,
and the object orientation in quaternion representation in addition to the
class probabilities.
This technique is inspired by PoseCNN~\citep{xiang2018posecnn}. Similar to PoseCNN,
we aggregate the center direction estimates in a Hough voting scheme. For each
detected object segment, the orientation estimates are aggregated using
quaternion averaging~\citep{markley2007averaging}.

\subsection{Pose Refinement using Rigid Registration}
\label{sec:pose_refinement}
Since the Kinect v2 sensor also provides depth, we can refine the pose estimation
using rigid registration against a suitable mesh. For this purpose, we choose
one \textit{canonical} mesh and perform Iterative Closest Point (ICP) registration of the measured and
segmented object points against it. This step largely helps in the depth component
of the object translation, since this is difficult to estimate from RGB alone.

This refinement step can also be used after grasping the object in order to measure
the object pose in-hand. In this case, points belonging to the hand are filtered out.
Instead of adapting the semantic segmentation for this purpose,
we opt to filter out a 3D cuboid around the end-effector position obtained from
forward kinematics. In practice, this ensures that there are no false positives
disturbing the refinement process.

\section{Regrasp Planning}
\label{sec:regrasp_planning}
Once the perception challenges have been resolved, i.e., the object is segmented and localized,
grasp planning methods are implemented to ensure the object's functional use.
First, a canonical model described by a mesh is deformed to match the observed object point cloud.
A functional grasping pose of the canonical model is correspondingly transformed to the observed model using the deformation field that deforms the mesh.
The resulting mesh is rendered from a top view and grasping candidates for the supportive arm are generated.
All grasping candidates are evaluated kinematically in combination with the functional grasp and a handover pose is generated.
The object is grasped according to the best candidate and brought to a dynamically determined observation position for in-hand object pose refinement.
Finally, the object is moved to the handover pose and the functional grasp is performed.

\subsection{Non-rigid (Shape Space) Registration}
\label{sec:shape_reg}
The main goal of the non-rigid registration is to deform a canonical model of the object category to fit the observed instance.
A category is defined as a group of objects with similar extrinsic geometry and usage, e.g., \textit{spray bottles} or \textit{watering cans}. 
The observation instance is represented as a point set---the result of the \textit{semantic segmentation} and \textit{pose estimation} module.
The canonical instance is represented as a triangular mesh.
The registration only acts upon the mesh vertex positions, so the connectivity and topology
of the mesh is preserved.

\begin{figure}[b!]
	\centering
	\includegraphics[height=1.6cm]{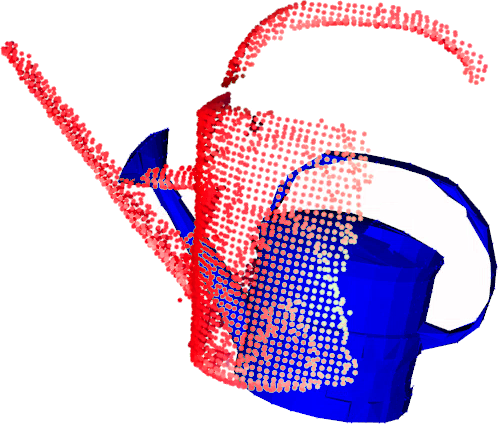}	
	\includegraphics[height=1.6cm]{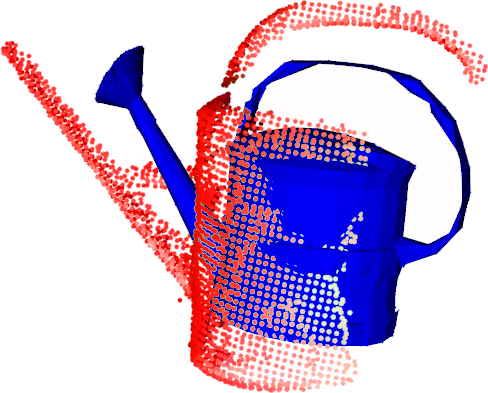}
	\includegraphics[height=1.6cm]{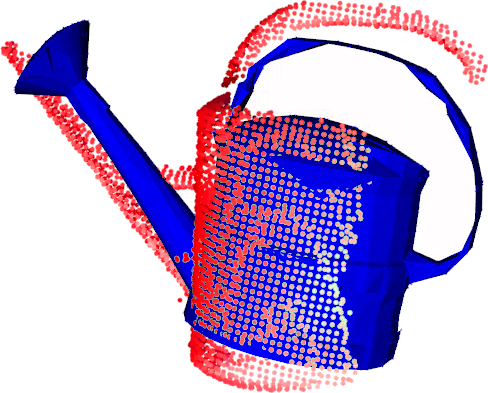}
	\includegraphics[height=1.6cm]{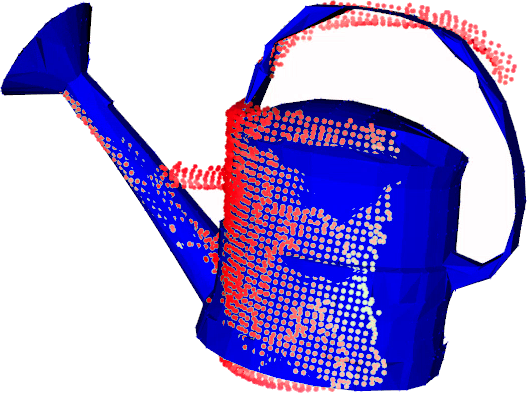}
	\caption{Shape registration. The canonical model (blue mesh) is deformed to match the observed object instance (red point cloud).
		The input of the registration is shown at the leftmost, while the result is presented at the rightmost.
		Note the scale difference and the robustness of the method against misalignment.}
	\label{fig:shape_reg}
\end{figure}

The point density of the mesh is a key factor for the performance of the non-rigid registration.
Larger faces may lead to poorly detailed registrations.
Therefore, linear subdivision filters are applied to the meshes to increase the point density by splitting larger faces.
The resulting high number of points is later reduced by quadric clustering filters to reduce computation time.

Our non-rigid registration embeds typical variations of an object category in a shape (latent) space.
The shape space can be interpreted as a regularization on the shape of the objects.
Shape space operations can be divided into the training phase (building the latent space)
and the inference phase.
The shape space is learned as a manifold spanned by the principal components of the deformation fields of one canonical model towards all the other instances of the training set.
The deformation fields are calculated by using the Coherent Point Drift (CPD).
In the inference phase, a shape space descriptor, i.e., a low-dimensional representation of a deformation field,
is found by searching in the shape space in a gradient descent manner, such that the deformed model matches the observation best according to a cost function.
This function includes the point distances to the closest points, a local rigid transformation to account for global misalignment and a regularization term for the latent variables~\cite{Rodriguez2018Learning}.
One advantage of using dense deformation fields is that points which do not belong to the canonical model can be deformed even after the registration has finished.
In this manner, grasping poses are transferred from the known canonical model to the novel instance.
For more in-depth discussion about the shape space registration please refer to~\citep{Rodriguez2018a} and~\cite{Rodriguez2018b}.

The shape space registration of a watering can is depicted in \cref{fig:shape_reg}.
The leftmost image shows the input of the registration,
i.e., the observing model (red points) and the canonical model (blue mesh).
Note the difference in size and robustness against misalignment.
The rightmost image shows the result of the registration which includes a global correction of the rigid transformation.

\subsection{Grasp Sampling of the Supportive Arm}
In order to obtain control over the object pose, an initial grasp is performed by the supportive arm.
Parts of the Dexterity Network (Dex-Net)\cite{dexnet1} pipeline are used to sample promising grasp hypotheses,
given the 6D object pose and the warped object mesh $M$.
Note that even though these initial grasps can be defined in the canonical model and transferred to the observed instance,
similar to the functional grasp (Sec. \ref{sec:shape_reg}),
this grasp hypotheses generation avoids any annotation effort at the cost of computation time,
less than \unit[4]{s} in our experiments (Table~\ref{table:runtime}).

Dex-Net uses a Multi-Armed Bandit model with correlated rewards for grasp planning.
Objects are classified by a Multi-View Convolutional Neural Network on a large data set using cloud computing.
This object classification defines a similarity metric between objects. 
Grasps are generated using an antipodal grasp sampler considering the gripper width and friction cones using a depth image and an object segmentation mask as input.
Since the grasp of the supportive arm depends on the functional one, 
only the Dex-Net's object sampling component and predicted prior belief distribution for each grasp are incorporated in our approach.

\begin{figure}
	\centering
	\includegraphics[height=3.2cm]{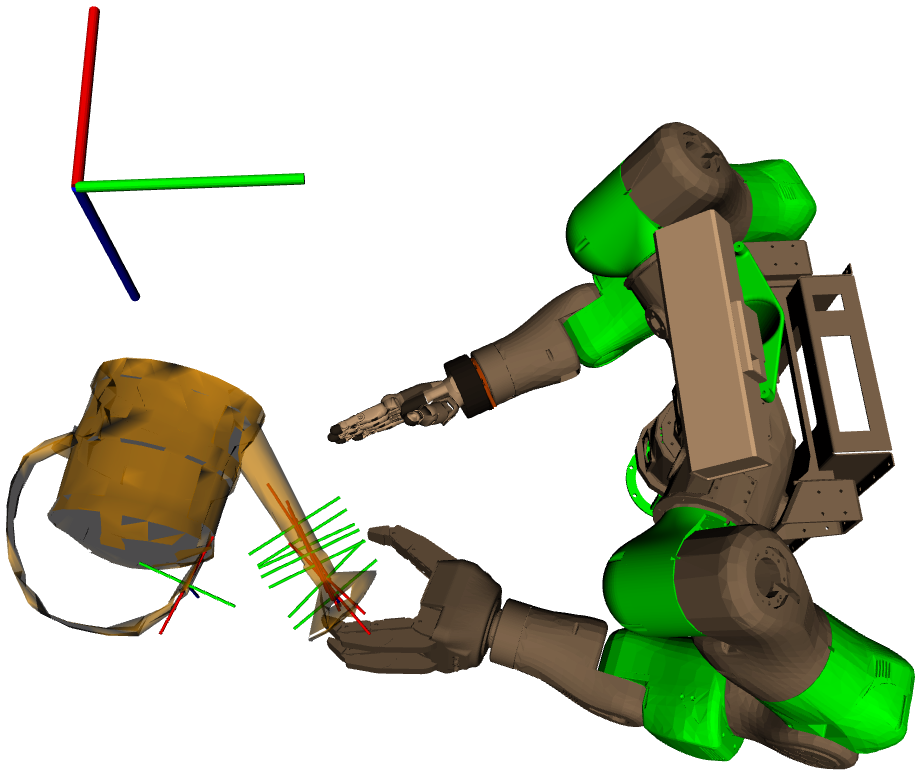}\hspace{1cm}
	\includegraphics[height=3.2cm]{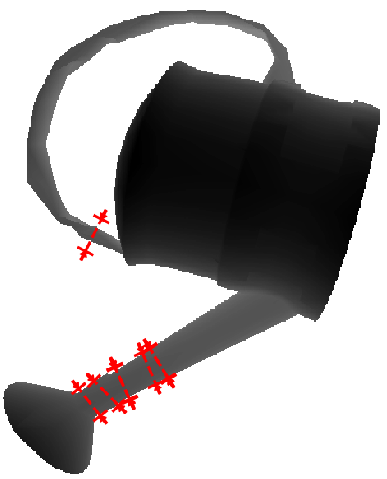}
	\caption{Grasp sampling of the supportive hand.
		Left: Virtual camera pose (largest axis), deformed object mesh and generated grasp samples (smaller axes).
		Right: Rendered depth image using the virtual camera containing generated grasp samples (red arrows).}
	\label{fig:sampler}
	\vspace*{-2ex}
\end{figure}

We generate a depth image $i_{\text{depth}}$ with a segmentation mask image $i_{\text{segm}}$,
both of them are passed to the Dex-Net pipeline and select the best grasp hypotheses according to their grasp success rate.
Since Dex-Net only generates grasps parallel to the input image plane,
we synthetically generate the images $I=(i_{\text{depth}},i_{\text{segm}})$ by rendering the deformed object mesh $M$ from a virtual camera pose.
The virtual camera is placed directly above the object center facing the object.
In this manner, grasps hypotheses are generated parallel to the surface under the object (e.g., a table)
which minimizes the chance of collisions between the gripper and the environment (see \cref{fig:sampler}). 
To increase the number of sampled grasp poses, 
multiple image pairs $I_i$ can be created with different virtual camera poses.

Each grasp hypothesis consists of the 3D grasp center position and direction, the gripper width and the estimated grasp success rate.
To determine the final set of candidate grasps $Q$, we greedily take the next best grasp,
while obeying a minimal horizontal separation of 1\,cm between the grasps.
Grasp selection stops when the estimated success rate surpasses a defined value but at least ten grasps have been selected.
Finally, the number of grasps are doubled by rotating each grasp by \ang{180} around the grasp axis.

The grasps hypotheses inferred by the network assume the usage of parallel grippers.
This assumption can be relaxed to multi-finger hands,
as presented here (Robotiq 3 Finger gripper),
by providing a grasping motion that resembles the one performed by a parallel gripper,
e.g., by thumb abducted grasps with anthropomorphic hands.

\subsection{Handover Pose Generation}
Given a set of grasps $Q$ for the supportive arm and a functional grasp $f$, 
we aim to find an optimal handover pose $h^*=(q^*, f^*)$, defined as a tuple of two 6D end-effector poses.
These poses are expressed relative to the base frame of the robot and assigned to each individual robot arm.
Note that all grasps contained in $Q$ and the functional grasp $f$ are defined for the initial position of the object, 
i.e., $f$ is not reachable due to collisions with environment or kinematics constraints,
thus the object has to be regrasped.

Initially, any grasp $q_i \in Q$, which is kinematically infeasible or causes self-collisions is filtered out.
Grasps, whose pregrasp pose is unfeasible due to similar reasons, are discarded as well.
The pregrasp pose for $q_i$ is defined by applying a fixed transform to $q_i$.
The main goal of the pregrasp pose is to ensure a desired approaching direction for the grasp.

The optimal handover pose $h^*$ is computed by minimizing a cost function $c$~(Eq. \ref{eq:proximity_cost}).
Rigid transformations $T_j$ are sampled to provide a handover configuration candidate $h_i^j$, when applied to $(q_i, f)$. 
To satisfy the closed kinematic chain,
the translation component of $T_j$ translates the end-effector poses in global frame, 
while the rotation component of $T_j$ makes the end-effector poses rotate with respect to the middle frame between them: $(q_i^t + f^t)/2$ using global frame axes,
whereas the superscript $t$ refers to the translation of the corresponding pose. 
Each computed handover candidate $h_i^j$ is then evaluated by the cost function $c$.
The handover pose computation is presented in Algorithm~\ref{alg:handover}.

\begin{algorithm}[t]
	\caption{Handover Pose Generation}
	\label{alg:handover}
	\textbf{Input:} Set of supportive grasps $Q$ and functional grasp $f$
	\begin{algorithmic}[1]
		\State $c_{min} = 1$; $c_{stop} = 0.1$
		\State $\forall q_i \in Q$:
		\begin{enumerate}
			\item Sample a set of transformations $T$.
			\item $\forall T_j \in T$:
			\begin{enumerate}
				\item Translate $(q_i, f)$ with translation component of $T_j$ in global frame.
				\item Rotate $(q_i, f)$ with rotation component of $T_j$ around midpoint $(q_i^t+f^t)/2$ using global  frame axes to obtain $h_i^j$.
				\item Check if handover candidate $h_i^j$ is kinematically feasible and collision-free. If not, proceed to the next $T_j$.
				\item Compute cost: $c_h = c(\mathcal{I}(h_i^j)) $~(Eq.~\ref{eq:proximity_cost}).
				\item If $c_h < c_{min}$: $c_{min}=c_h$; $h^*=h_i^j$.
			\end{enumerate}
			\item Early stop: if $c_{min} < c_{stop}$: break.
		\end{enumerate}
	\end{algorithmic}
	\textbf{Output:} Optimal handover configuration $h^*=(q^*, f^*)$ with minimal cost.
\end{algorithm}

The cost function $c$ is defined as a joint limits proximity cost which penalizes closeness to joint limits.
In this manner, we aim to be far from stretched configurations, which generally result in kinematic singularities.
Given joint positions $\bm{\theta}$ (calculated from $h_i^j$ by means of the inverse kinematics $\mathcal{I}$),
upper and lower joint limits $\bm{\theta}_{upper}$ and $\bm{\theta}_{lower}$, 
joint limit proximities $\delta(\bm{\theta})$ are computed as:
\begin{equation}
\delta(\bm{\theta}) = \min(|\bm{\theta}_{upper}-\bm{\theta}|, |\bm{\theta}-\bm{\theta}_{lower}|),
\label{eq:proximities}
\end{equation}
which define the joint proximity cost as:
\begin{equation}
c(\bm{\theta}) = 
\frac{1}{|\delta(\bm{\theta})|} \sum_{i=1}^{|\delta(\bm{\theta})|} \frac{1}{\varepsilon^2}(\delta(\bm{\theta}_i))^2 - \frac{2}{\varepsilon} \delta(\bm{\theta}_i) + 1,
\label{eq:proximity_cost}
\end{equation}
where $\varepsilon$ is a constant which represents the maximum proximity to the joint limits.
The cost function is close to $1.0$ when approaching the proximity of $0$. 
Moreover, the value range of the cost function is $[0, 1]$, which makes it easy to integrate with additional cost components, if necessary.

To perform fast collision checking we split the world representation into two parts: static and dynamic, similar as in~\cite{Pavlichenko2017}. 
The static part contains the environment and the non-moving robot bodies and is represented by means of Euclidean Distance Field (EDT).
Collision approximation (e.g., geometric primitives) of the robot body is known in advance. 
Collision information about the environment can be added to the EDT online using point clouds. 
Finally, the robot dynamic part represents the moving robot bodies, namely the arms and the hands--- approximated by a set of spheres. 
The object held in the hand is approximated by a bounding sphere as well.

\subsection{View Pose Generation}
Once we have the handover configuration $h^*=(q^*, f^*)$, 
where $q^*$ is the grasp of the supportive hand and $f^*$ is the functional grasp pose for the second hand, the first grasp is executed. 
During the grasp, the object may move arbitrarily.
This movement is very hard to predict. 
Moreover, even a slight change in the object pose estimate can lead to the failure of the functional grasp, which requires high precision to succeed. 
Thus, in-hand pose estimation is performed once the object is grasped.
In this manner, a displacement $T_{view}$ between the expected object pose in hand and the actual object pose can be calculated.
By applying this displacement, the functional grasp is refined $f' = f^* T_{view}$.
The displacement $T_{view}$ is calculated as explained in \cref{sec:pose_refinement}.

Once the first grasp is executed, 
there is no guarantee that enough points on the object surface are observable for the pose refinement, mainly because the arm and the hand may obstruct the sensor's field of view. For this reason, an observation pose $f_o$ is computed to increase the number of by-the-sensor observable points.

The view pose generation is agnostic of the object and of the handover pose computation. 
Since we want to functionally grasp the object, 
we can assume the presence of observable geometric features in a region around $f^*$.
The minimal distance for obtaining quality sensor readings is $d_{min} = 0.5$m for the Kinect v2, 
moreover, placing the object in the center of the image will decrease radial distortion and reduce the chances of the object being cropped. 
Thus, we define a pose $f_o$ by translating the camera frame along its Z axis by $d = d_{min} + D$, where $D = 0.1$m is a predefined offset. 
Regarding the orientation of $f_o$, the approaching vector, defined as a vector from the pregrasp to the grasp pose is oriented parallel to the Z-axis of the sensor.

However, in many situations, it will not be possible to achieve $f_o$ due to the robot kinematic restrictions or collisions. 
To solve that, we sample poses around $f_o$ and select the pose which is kinematically feasible and is the closest to the canonical pose $f_o$.

\section{Evaluation}
\label{sec:evaluation}
\begin{figure}
	\centering
	\includegraphics[width=0.9\linewidth]{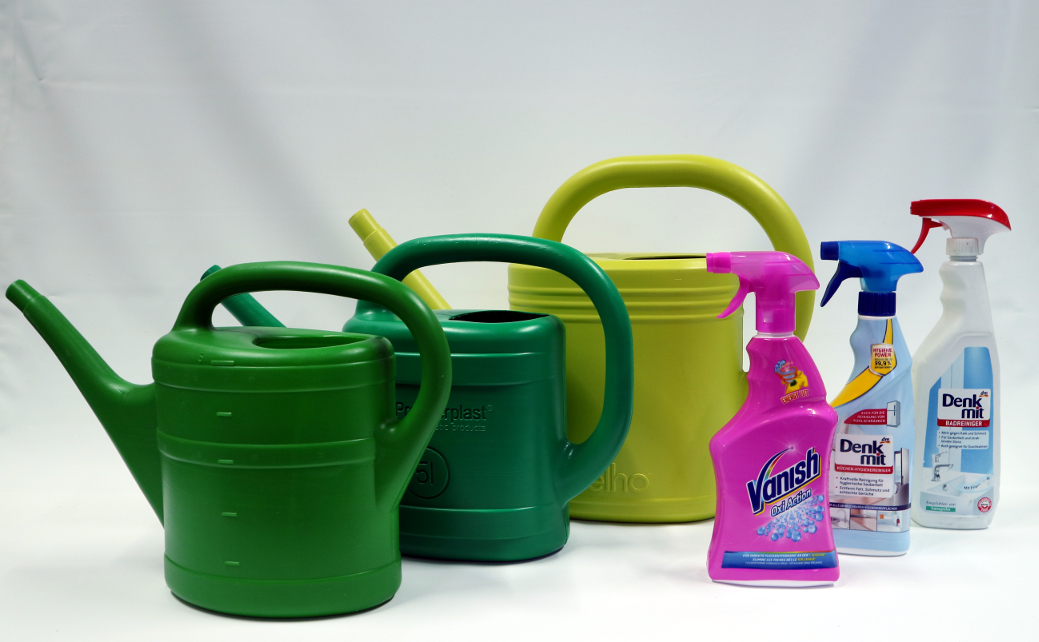}
	\caption{Three watering cans and three spray bottles used during the evaluation. Please note that none of the instances was observed by the robot before the evaluation.}
	\label{fig:all_objects}
	\vspace*{-2ex}
\end{figure}

We evaluate the proposed system on the real humanoid upper body introduced in
\cref{sec:system_overview}. The robot is mounted in front of the table, on which the objects are placed. The task is to grasp the object with the left hand and perform a successful handover to the right hand, so that the desired functional grasp is achieved. The objects are lying on the table in a way that the desired functional grasp cannot be achieved directly because it is kinematically impossible or due to collisions with the table.
We evaluate the system on three previously unseen different instances of the following two object categories: \textit{spray bottle} and \textit{watering can} (\cref{fig:all_objects}).

The shape space of the \textit{spray bottle} and \textit{watering can} categories were built using 16 and 8 meshes respectively.
The meshes were obtained from 3D online databases\footnote{\url{3dwarehouse.sketchup.com}, \url{https://grabcad.com} and \url{https://sketchfab.com}} and were preprocessed as explained in Sec.\ref{sec:shape_reg}.
All the meshes were manually aligned to a common object frame.
The canonical models were selected by experts.

For training the semantic segmentation and pose estimation network, we generate
synthetic scenes using the Magnum 3D rendering
engine\footnote{\url{https://magnum.graphics}}.
We sample background images from nine images captured by the robot (with no objects in the
workspace). To increase robustness, we also sample images from the ObjectNet3D
dataset~\citep{xiang2016objectnet3d} as backgrounds with a 50\% probability.
The backgrounds are randomly cropped and scaled to a resolution of $960 \times 540$.
We then render one to five random object meshes on top, with uniformly drawn
orientations and translations s.t. the object center is in the camera frustum.
The rendering pipeline generates the required ground truth for training
(i.e. segmentation information) at the same time.
The network is trained using the Adam optimization technique with a learning
rate of $1\cdot 10^{-4}$ for 120k iterations.
We train separate network models for watering cans and spray bottles.

For generating and evaluating supportive grasp candidates, 
we used the Dex-Net\footnote{\url{https://berkeleyautomation.github.io/dex-net/}} architecture without training new CNN models. 
We fine-tuned some of the internal sampling parameters and adapted the grasp width for each object category.

For each of three spray bottles we perform three handover attempts for three different initial orientations of the spray bottle. We proceed as follows: the bottle lies on its side on the table in front of the robot. The functional grasp is not achievable directly. The yaw of the orientation of the bottle is: \ang{-45}/\ang{0}/\ang{+45}, where \ang{0} corresponds to the bottle being perpendicular to the robot torso. This results in 27 functional regrasping attempts. For watering cans we performed a total of 26 attempts. The watering cans were lying on a side in a similar manner. However, due to the object size and arm reach we tested a smaller range of orientations. The resulting success rates are shown in Table~\ref{table:success_rate}.

\begin{table}
	\centering
	\caption{Regrasping success rate}
	\label{table:success_rate}
	\begin{tabular}{ccccc}
		\toprule
		Category & Object & Func. Grasp & Grasp & Trials \\
		\midrule
		\multirow{5}{*}{Spray Bottle} & White & 8 & 9 & 9 \\
		&Blue & 6 & 7 & 9\\
		&Pink & 5 & 8 & 9\\
		\cmidrule{2-5}
		&\textbf{Overall} & \textbf{19 (70\%)} & \textbf{24 (88\%)} & \textbf{27}\\
		\midrule
		\multirow{5}{*}{Watering Can} &Dark green& 4 & 4 & 6 \\
		&Light green& 6 & 6 & 10\\
		&Yellow& 7 & 7 & 10\\
		\cmidrule{2-5}
		&\textbf{Overall} & \textbf{17 (65\%)} & \textbf{17 (65\%)} & \textbf{26}\\
		\bottomrule
	\end{tabular}
	\vspace*{-3ex}
\end{table}

\begin{figure}[b!]
	\vspace*{-1ex}
	\centering
	\includegraphics[width=4.cm]{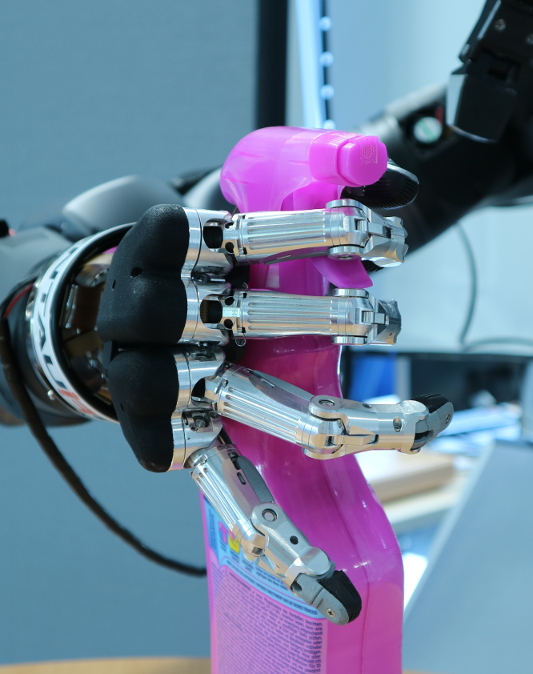}\hfill
	\includegraphics[width=4.cm]{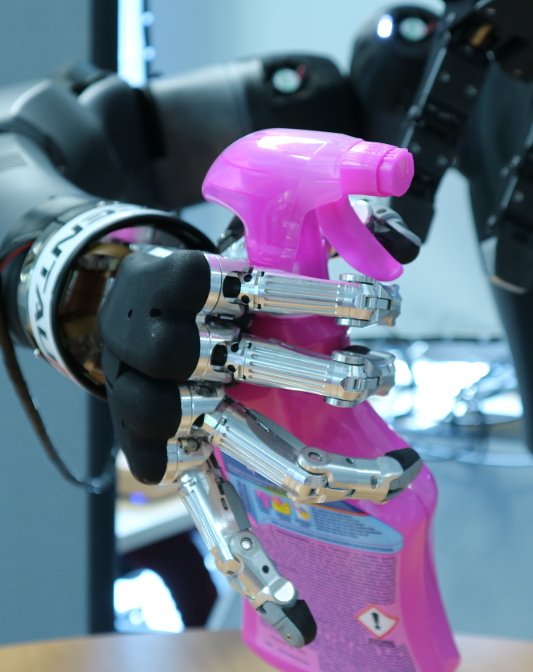}
	\caption{Example of a successful (left) and failed functional grasp (right).}
	\label{fig:grasps}
\end{figure}

It is possible to see that for both categories the success rate is similar: 65~\% for watering cans and 70~\% for spray bottles. In case of spray bottles it happened relatively frequently (18~\%)) that the object was handed over successfully but the resulting grasp was not functional. This happened because a functional grasp for a spray bottle is a very specific pose, where the pointing finger is at the trigger. Consequently, even a very small error in any of the components could lead to a failure. Often we had a case when the finger would miss the trigger. The cause for this was mostly the imperfect mesh registration or slight mismatch in the correction step during ICP. For the three cases of a complete failure, the causes were as follows: in two cases the ICP during the correction was misaligned by \ang{90}, in one case a kinematically possible handover pose was not found after the correction step. Examples of a successful and failed functional grasps are shown in \cref{fig:grasps}. For the watering cans the major issue was the supportive grasp: in six cases the supportive grasp was not reliable because it was applied to the tip of the nose of the can, thus, one of the fingers was not making a firm contact with the object. This can be solved by additional validation of grasp candidate quality during the supportive grasp sampling or prior to handover pose computation. In another case the supportive grasp failed because the fingers collided with an extended wide part of the nose, which was not modeled by mesh registration. Finally, in two cases the final grasp with the right hand was missed due to slight misalignment in ICP during the correction step.

\captionsetup[subfigure]{labelformat=empty}
\begin{figure*}[bt!]
	\centering
	\subfloat[(a)]{\includegraphics[width=3.4cm]{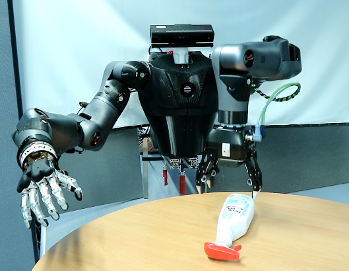}}\hfill
	\subfloat[(b)]{\includegraphics[width=3.4cm]{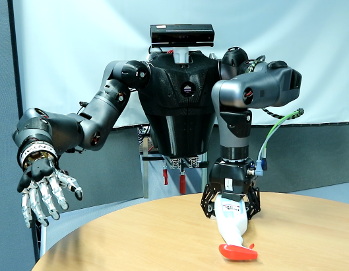}}\hfill
	\subfloat[(c)]{\includegraphics[width=3.4cm]{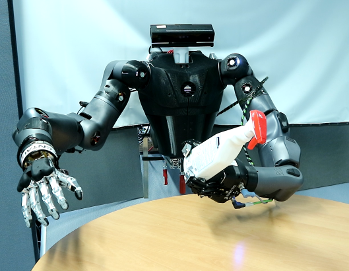}}\hfill
	\subfloat[(d)]{\includegraphics[width=3.4cm]{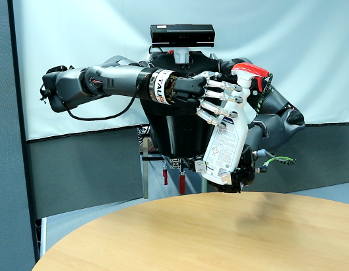}}\hfill
	\subfloat[(e)]{\includegraphics[width=3.4cm]{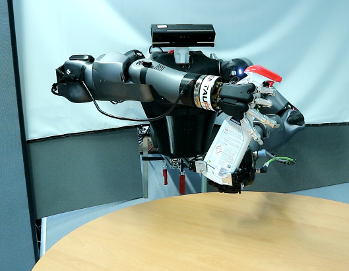}}
	\\
	\subfloat[(a)]{\includegraphics[width=3.4cm]{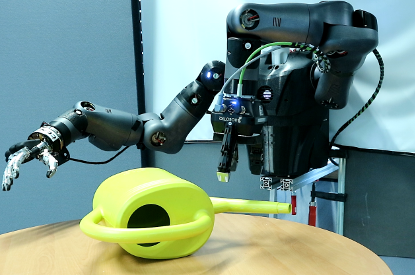}}\hfill
	\subfloat[(b)]{\includegraphics[width=3.4cm]{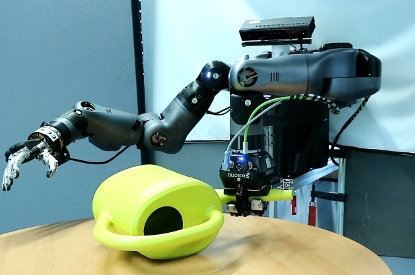}}\hfill
	\subfloat[(c)]{\includegraphics[width=3.4cm]{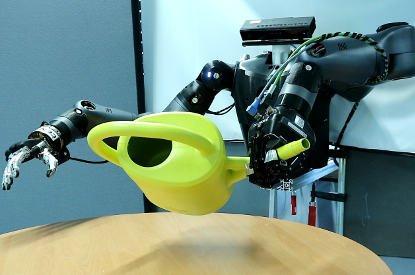}}\hfill
	\subfloat[(d)]{\includegraphics[width=3.4cm]{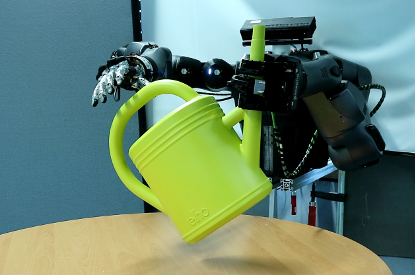}}\hfill
	\subfloat[(e)]{\includegraphics[width=3.4cm]{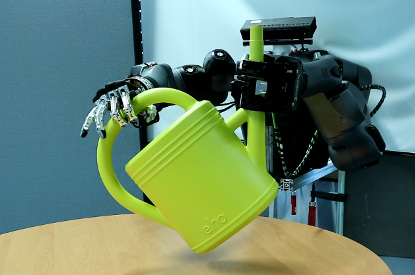}}
	\caption{Dual-arm functional regrasp of a spray bottle and a watering can. \textbf{(a)} Pregrasp before the supportive grasp. \textbf{(b)} Supportive grasp. Note how both objects changed their pose. \textbf{(c)} Observation pose. In-hand object pose estimation is performed to correct for the disturbance caused by the supportive grasp. \textbf{(d)} - \textbf{(e)} Functional grasp.}
	\label{fig:real_robot_regrasp}
	\vspace*{-2ex}
\end{figure*}

We measured the runtime of each of the components involved into the pipeline. The resulting runtimes are shown in \cref{table:runtime}. One can observe that for the watering cans category the runtime of the pipeline is almost two times bigger than runtime for the spray bottles category. This is because the watering cans are much larger than the spray bottles, and, hence, the observations contain much more points for this category. It severely affects the mesh registration as well as ICP during the in-hand object pose correction. The computation time for EDT used for collision checking is not mentioned, because it can run in parallel with other components of the pipeline. On average, the EDT computation took 1.27 $\pm$ 0.37\,s.

Overall, the performed evaluation demonstrated that the proposed pipeline is capable of performing the functional regrasping completely autonomously in real-world scenarios. An example of a functional regrasp of a spray bottle and a watering can is shown in \cref{fig:real_robot_regrasp}. The system demonstrated a success rate of ~70\% for two object categories, maintaining runtime of 15-30\,s which allows for executing such tasks online. The video of the experiments is available online\footnote{Experiment video:~\url{http://www.ais.uni-bonn.de/videos/Humanoids_2019_Bimanual_Regrasping}}.

\begin{table}[]
	\centering
	\caption{Runtime [s] per pipeline component}
	\label{table:runtime}
	\begin{tabular}{lr@{ $\pm$ }lr@{ $\pm$ }l}
		\toprule
		Component                    & \multicolumn{2}{c}{Spray bottles} & \multicolumn{2}{c}{Watering cans} \\
		\midrule
		Semantic Segmentation        & 0.74 & 0.02 & 0.74  & 0.02 \\
		Pose Estimation + ICP        & 2.20 & 0.04 & 2.20  & 0.04  \\
		Mesh Registration            & 8.59 & 0.88 & 15.83 & 0.60 \\
		Supportive Grasp Sampling    & 2.47 & 0.13 & 3.24  & 0.06 \\
		Handover Pose Computation    & 0.81 & 0.43 & 1.75  & 0.50  \\
		Observation Pose Computation & 0.07 & 0.01 & 0.07  & 0.01 \\
		In-Hand Pose Correction (ICP)& 2.09 & 0.19 & 5.17  & 0.88 \\
		\midrule
		Complete Pipeline            & 16.97 & 1.70 & 29.00 & 2.11  \\
		\bottomrule		
	\end{tabular}
	\vspace*{-2ex}
\end{table}

\section{Conclusions}
\label{sec:conclusion}
We proposed a complete integrated pipeline for autonomous bimanual functional regrasping of previously unseen objects of a known category. The approach utilizes a single sensor --- Kinect v2 --- to solve the task. The pipeline consists of: semantic segmentation and 6D object pose estimation, non-rigid model registration, supportive arm grasp candidate generation, and handover pose planning. To mitigate the imperfection of the supportive grasp, which may cause the object to move in an unpredictable manner, we perform an additional in-hand object pose estimation to correct the initial estimated pose.

We perform experiments on a human-like upper body equipped with two 7 DOF arms with a Schunk Hand and a Robotiq 3F Gripper as end-effectors. The system is tested on six previously unseen objects of two categories: spray bottles and watering cans. The pipeline achieved a success rate of 70~\% and 65~\% per category, with computations done in 15-30~s. For the spray bottles category the main source of failures are slight errors during the in-hand pose estimation, since this category requires a very precise functional grasp. In contrast, for the watering cans, which are represented by bulkier objects with complex shape, but requiring a less precise functional grasp, the main source of failures was an unstable supportive grasp. The described evaluation showed that our pipeline can be successfully applied to solve real-world functional regrasping problems.

In the future, this approach can be extended to incorporate multiple handover actions in cases where a single handover is not enough to enable a functional regrasping.
Another open problem is the use of environmental constraints, e.g., by placing the object on a surface for a direct functional grasp or for a handover procedure.

\balance
\bibliographystyle{IEEEtranN}
\bibliography{regrasping}

\end{document}